\documentclass{article}

\usepackage[nonatbib, final]{neurips_2021}

\usepackage[utf8]{inputenc} 
\usepackage[T1]{fontenc}    
\usepackage{hyperref}       
\usepackage{url}            
\usepackage{booktabs}       
\usepackage{amsfonts}       
\usepackage{nicefrac}       
\usepackage{microtype}      
\usepackage{xcolor}         
\usepackage{graphicx}
\usepackage{tabulary}
\usepackage{multirow}
\usepackage{pifont}
\usepackage{url}
\usepackage{xcolor}
\usepackage[super]{nth}
\usepackage{caption}
\usepackage{subcaption}
\usepackage{float}
\usepackage[utf8]{inputenc}
\usepackage[english]{babel}
\usepackage{amsmath}
\usepackage{hyperref}
\usepackage{multicol}
\usepackage{bbold}
\usepackage{amsfonts}
\usepackage{cite}
\usepackage[export]{adjustbox}
\usepackage{ragged2e} 
\usepackage{mathtools}
\usepackage{hyperref}
\usepackage{amsthm}
\newcommand{\defeq}{\vcentcolon=}

\newtheorem{theorem}{Theorem}
\newtheorem{lemma}[theorem]{Lemma}

\usepackage{wrapfig}

\title{Conditional Alignment and Uniformity for Contrastive Learning with Continuous Proxy Labels}

\author{Benoit Dufumier$^{1,2}$\thanks{Corresponding author} \quad Pietro Gori$^{2}$\quad Julie Victor$^{1}$\quad Antoine Grigis$^{1}$\quad Edouard Duchesnay$^{1}$ \\ $^1$NeuroSpin, CEA Saclay, Université Paris-Saclay, France \\ $^2$LTCI, Télécom Paris, IPParis, France\\ \texttt{benoit.dufumier@cea.fr} }
  
\begin{document}

\maketitle

\begin{abstract}
  Contrastive Learning has shown impressive results on natural and medical images, without requiring annotated data. However, a particularity of medical images is the availability of meta-data (such as age or sex) that can be exploited for learning representations. Here, we show that the recently proposed contrastive $y$-Aware InfoNCE loss, that integrates multi-dimensional meta-data, asymptotically optimizes two properties: \textit{conditional alignment} and \textit{global uniformity}. Similarly to \cite{wang2020}, conditional alignment means that similar samples should have similar features, but conditionally on the meta-data. Instead, global uniformity means that the (normalized) features should be uniformly distributed on the unit hyper-sphere, independently of the meta-data. 
  Here, we propose to define \textit{conditional uniformity}, relying on the meta-data, that repel only samples with dissimilar meta-data. We show that direct optimization of both conditional alignment and uniformity improves the representations, in terms of linear evaluation, on both CIFAR-100 and a brain MRI dataset.
\end{abstract}

\section{Introduction}

Deep learning models have recently shown impressive results in medical imaging \cite{azizi2021big, gulshan2016retino,mckinney2020breast,esteva2017dermatologist}, especially when large data-sets are available. However, in many medical applications (\textit{e.g}, computer-aided diagnosis with MRI \cite{wen2020}, chest X-ray on COVID-19 \cite{sowrirajan2021moco}, etc.), data-sets are often 
small or heterogeneous and biased \cite{tartaglione2021end} (\textit{e.g}, images coming from different hospitals) and annotations are costly.

Traditional transfer learning from natural images has been applied to several medical applications \cite{alzubaidi2020, sahlsten2019}, hypothesizing that 
features can be re-used or fine-tuned on the downstream tasks \cite{yosinski2014transferable}. However, it does not always lead to better results \cite{raghu2019transfusion}, which might be due to the large domain gap between natural and medical images. On the contrary, self-supervised models do not rely on annotations and can be directly applied to big unlabeled medical data-sets, avoiding both the burden of expert annotations and the domain gap. These models are trained on a pretext task that incorporates prior information about the data and should estimate 
a relevant representation for the subsequent downstream tasks (\textit{e.g}, the invariants \cite{chen2020simCLR}, object shapes or colors \cite{noroozi2016jigsaw, larsson2017colorproxy, pathak2016inpainting}, etc). Recent approaches are built upon Siamese networks \cite{bromley1993siamese}, as in contrastive learning \cite{chen2020simCLR, he2020moco}, where 
an encoder is trained to map different views of the same image to the same point in the representation space while pushing away dissimilar images. In practice, the InfoNCE loss \cite{gutmann2010infoNCE, oord2018representation, chen2020simCLR} is 
usually used and it has shown impressive results on both natural \cite{chen2020simCLR, he2020moco} and medical \cite{azizi2021big, taleb20203d, chaitanya2020contrastive} image classification and segmentation. 

A particularity with medical images is the availability of meta-data (such as participant's age and sex) that often come freely with medical data-sets and that can be viewed as prior knowledge. As observed in \cite{chaitanya2020contrastive, dufumier2021yaware}, most of the methods in contrastive learning consider distinct images \textit{equally}  semantically different. However, two different images with close meta-data should probably share discriminative features, as opposed to images with very distinct meta-data. In \cite{dufumier2021yaware}, a new contrastive loss based on InfoNCE and integrating continuous meta-data has been proposed, namely $y$-Aware InfoNCE. It generalizes the SupCon loss \cite{khosla2020} and has shown good results on brain MRI. We shall see its connection with the alignment and uniformity terms proposed in \cite{wang2020}.

\begin{wrapfigure}{L}{0.5\textwidth}
  \vspace{-10pt}
  \centering
    \includegraphics[width=0.5\textwidth]{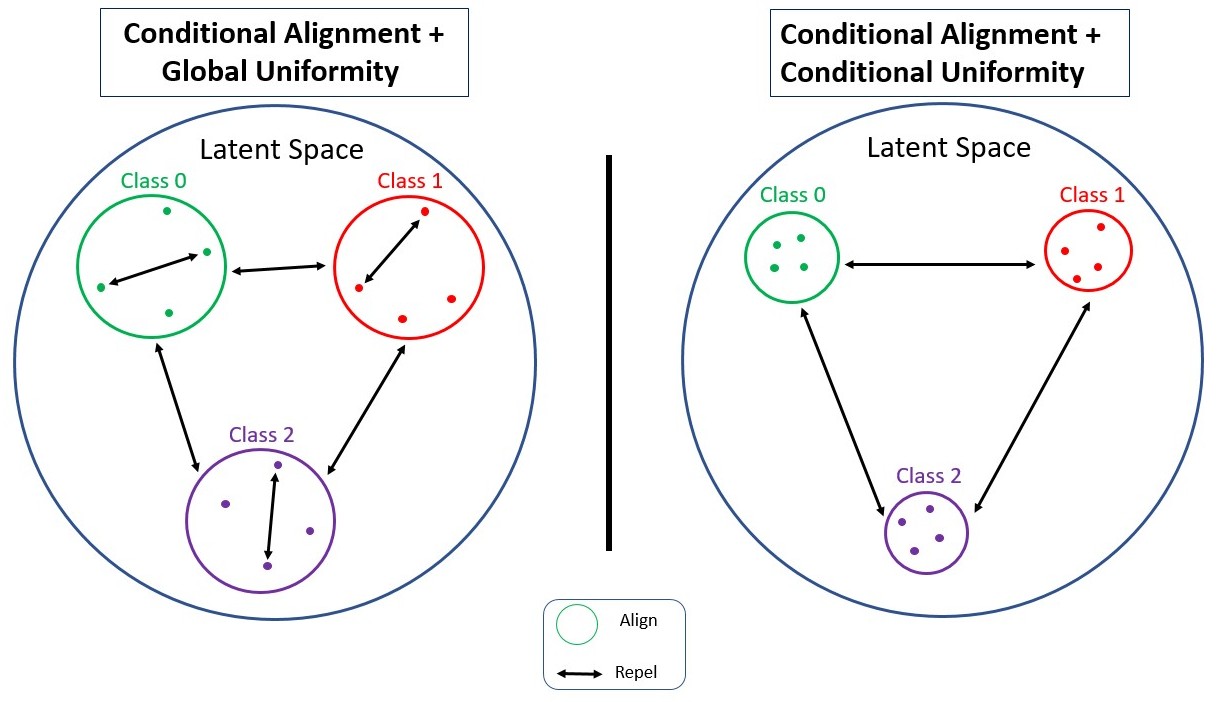}
  \caption{Proposed conditional alignment and uniformity terms with categorical meta-data. The conditional uniformity repel only points from different classes and not within the same class.}
  \label{fig:global_conditional_uniformity}
  \vspace{-8pt}
\end{wrapfigure}

\textbf{Contributions.} First, we show that the $y$-Aware InfoNCE loss can be decomposed, asymptotically, into \textit{conditional} alignment (pulling images with close meta-data $y$ close together) and \textit{global} uniformity (repelling all images independently of $y$, see Fig. \ref{fig:global_conditional_uniformity}). Second, remarking that images should not be uniformly repelled, and inspired by \cite{chuang2020debiased}, we propose to re-define the uniformity term only between samples with dissimilar meta-data (ideally coming from different latent classes), calling it conditional uniformity. We empirically demonstrate that going from global to conditional uniformity allows the encoder to learn a better transferable representation on both CIFAR-100 and a brain MRI dataset.  

\section{Problem Formulation and $y$-Aware InfoNCE estimator}

In contrastive learning, the goal is to learn an embedding $f_\theta: \mathcal{X}\mapsto \mathcal{S}^{d}$ that maps similar samples $(x, x^+) \in \mathcal{X}\times \mathcal{X}$ to close representations $f_\theta(x), f_\theta(x^+)$ but also different samples $(x, x^-) \in \mathcal{X}\times \mathcal{X}$ to distant representations $f_\theta(x), f_\theta(x^-)$, on a unit hyper-sphere $\mathcal{S}^d$. Similarly to \cite{dufumier2021yaware}, we assume that each input sample $x$ has a proxy label $y\in \mathbb{R}^p$ giving prior knowledge about $x$ (such as participant's age and/or sex) and coming from the joint distribution $p(x, y)$. 

For an anchor $(x, y)\sim p(x,y)$ and $N$ other samples $(x_k, y_k)\sim p(x,y)$, the $y$-Aware InfoNCE \cite{dufumier2021yaware} is defined as:
\begin{equation}
    \mathcal{L}^y_{NCE} = - \frac{1}{N} \sum_{k=1}^N \frac{w_\sigma(y, y_k)}{\hat{Z}_\sigma(y)} \log \frac{e^{f_\theta(x, x_k)}}{\frac{1}{N}\sum_{j=1}^N e^{f_\theta(x, x_j)}}
\end{equation}

where $\hat{Z}_\sigma(y)=\frac{1}{N}\sum_{j=1}^N w_\sigma(y, y_j)$, $f_\theta(x_1, x_2)=\frac{1}{\tau}f_\theta(x_1)^Tf_\theta(x_2)$ and $w_\sigma$ is a Gaussian kernel that measures similarity between the proxy labels $y$. Similarly to \cite{wang2020}, this loss can be re-written as:
\begin{equation}
    \mathcal{L}^y_{NCE} = \underbrace{-\frac{1}{N}\sum_{k=1}^N \frac{w_\sigma(y, y_k)}{\hat{Z}_\sigma(y)} f_\theta(x, x_k)}_{\text{Conditional Alignment}} + \underbrace{\log \frac{1}{N}\sum_{j=1}^N e^{f_\theta(x, x_j)}}_{\text{Global Uniformity}}
\end{equation}

\begin{lemma}
    As the number of samples $N\rightarrow \infty$, the $y$-Aware InfoNCE loss $\mathcal{L}^y_{NCE}$ converges to:
    \begin{equation}
        \lim_{N\rightarrow\infty} \mathcal{L}^y_{NCE} = -\underbrace{\mathbb{E}_{p_{pos}(x, x^+)}(f_\theta(x,x^+))}_{\mathcal{L}_{align}^y} + \underbrace{\mathbb{E}_{x\sim p(x)}\log\mathbb{E}_{x'\sim p(x')}(e^{f_\theta(x, x')})}_{\mathcal{L}_{unif}}
    \end{equation}
    where $p_{pos}(x, x^+) = \int p(x|y) p(x^+|y^+) p_y^+(y^+)p(y)dydy^+ $ is a positive distribution satisfying $p_{pos}(x)=p(x)$ and $p_y^+(y^+) = \frac{1}{Z_\sigma(y)}w_\sigma(y, y^+)p(y^+)$ quantifies the similarity between $y$ and $y^+$. $Z_\sigma(y)=\mathbb{E}_{p(y^+)}(w_\sigma(y, y^+))$ is a normalizing constant. See the Appendix for a proof.
\end{lemma}


Similarly to \cite{wang2020}, the alignment term brings all positive pairs $(x, x^+)$ close together in $\mathcal{S}^d$ while the uniformity term is optimal when all points $f_\theta(x)$ for $x\sim p(x)$ are uniformly distributed on the hyper-sphere. In our formulation, we know this is hardly true since all points with close meta-data (measured by $p_y^+(y^+)$) should be close in $\mathcal{S}^d$. Instead of repelling uniformly all points from one another, we propose to \textit{weight} the repulsion between points with the meta-data, as in the conditional alignment. To this end, we introduce a negative distribution $p_{neg}(x, x^-)$ (see below). 


\paragraph{Conditional Uniformity.} As proved in \cite{wang2020}, the minimizers $f_\theta$ of $\mathcal{L}_{unif}$ minimize $\log \mathbb{E}_{x, x'\sim p(x)p(x')}(e^{f_\theta(x, x')})$. Instead of sampling independently $x, x'$, we propose to draw them from $p_{neg}(x, x^-) \defeq \int p(x|y)p(x^-|y^-)p_y^-(y^-)p(y)dydy^-$ where $p_y^-(y^-)=\frac{||w_\sigma||_\infty-w_\sigma(y, y^-)}{||w_\sigma||_\infty-Z_\sigma(y)}p(y^-)$. The infinite norm ensures the positiveness of our distribution and the kernel $w_\sigma(y, y^-)$ makes a direct connection with $p^+_y(y^+)$. We then define:
\begin{equation}
    \mathcal{L}_{unif}^y = \log \mathbb{E}_{(x, x^-)\sim p_{neg}(x, x^-)}(e^{f_\theta(x, x^-)})
\end{equation}
that can be empirically estimated using: $\hat{\mathcal{L}}_{unif}^y=\log \frac{1}{N^2}\sum_{i,j}\frac{||w_\sigma||_\infty-w_\sigma(y_i, y_j)}{||w_\sigma||_\infty-\hat{Z}_\sigma(y_i)}e^{f_\theta(x_i, x_j)}$.

\section{Experiments and Results}
\vspace{-10pt}


\begin{wrapfigure}{L}{0.62\textwidth}
  \vspace{-10pt}
  \centering
    \includegraphics[width=0.62\textwidth]{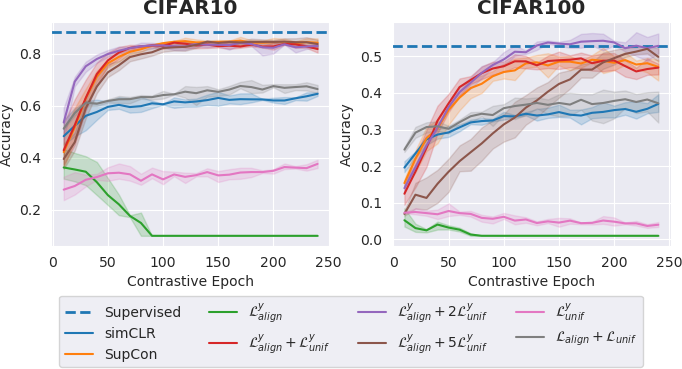}
  \caption{Linear Evaluation of ResNet18 pre-trained on CIFAR 10 and 100.}
 \label{fig:cifar_linear_eval}
 \vspace{-15pt}
\end{wrapfigure}

\textbf{CIFAR.} We pre-trained a ResNet18 \cite{ResNet_He} on CIFAR 10 and 100 \cite{krizhevsky2009cifar} by considering the true labels as meta-data and we compared our approach with different baselines: SimCLR  (unsupervised)\cite{chen2020simCLR}, SupCon (supervised with labels)\cite{khosla2020} and Alignment and Uniformity (unsupervised)\cite{wang2020} (see Appendix for experimental details). We reported in Fig. \ref{fig:cifar_linear_eval} the representation quality under the linear evaluation protocol. We observed a better convergence speed on CIFAR10 and a better representation on CIFAR100.

\begin{wrapfigure}{R}{0.62\textwidth}
  \vspace{-10pt}
  \centering
    \includegraphics[width=0.62\textwidth]{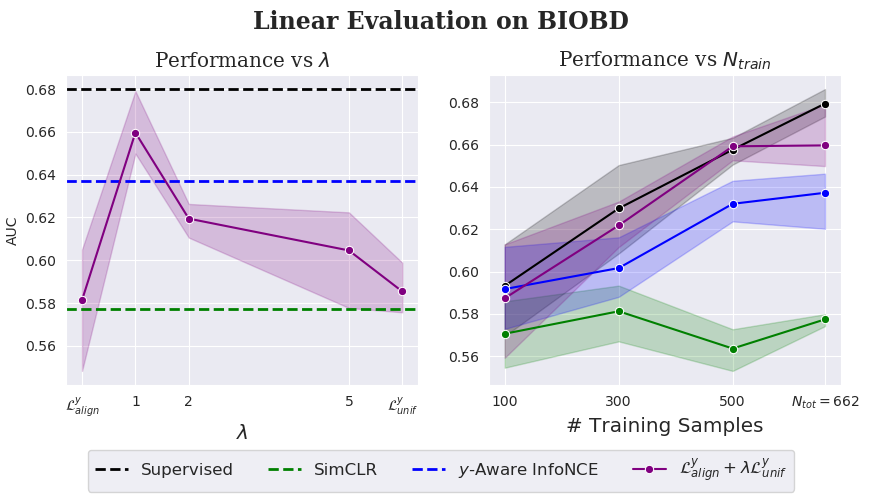}
  \caption{Linear Evaluation on BIOBD after pre-training on BHB-10K, varying $\lambda$ (left image with $N_{train}=662$) and the number of training samples (right image with $\lambda=1$).}
  \label{fig:linear_eval_biobd}
  \vspace{-35pt}
\end{wrapfigure}

\textbf{Brain MRI.} In a real case scenario, following the experimental setup in \cite{dufumier2021yaware}, we pre-trained a DenseNet121 \cite{DenseNet_Huang} on BHB-10K, a large multi-site brain MRI dataset comprising $10^4$ 3D scans of healthy controls. We used age and sex as meta-data with the associated kernel $\mathbf{w}_\sigma((age_1,sex_1),(age_2,sex_2))=w_\sigma(age_1, age_2)\delta_{sex_1=sex_2}$. We then assessed the representation quality on BIOBD \cite{hozer2020biobd, sarrazin2018}, to discriminate between bipolar disorder patients and healthy controls. 


\section{Conclusion}

By decomposing the recently proposed $y$-Aware InfoNCE loss into conditional alignment and global uniformity, we established a connection with the framework recently proposed in \cite{wang2020}. We also demonstrated that a conditional uniformity term is preferable and lead to a better representation quality on both CIFAR 100 and a brain MRI dataset. Future work will consist in validating the proposed framework on various modalities (\textit{e.g} CT and X-ray) and applying transfer learning from the learnt representations. 

\section{Broader Impact}
Including continuous and categorical meta-data into a contrastive loss should enrich the learnt representation making it more discriminative for any downstream task. This should particularly benefit the applications with small data-sets such as rare diseases or private clinical data-sets. Furthermore, the proposed loss could be also adapted for multi-modal data or for data-sets with missing imaging data, thus increasing and broadening its clinical impact and applicability. 

\bibliographystyle{plain}
\bibliography{bibliography}
\newpage
\appendix

\section{Experimental Details}

\paragraph{CIFAR-10 and 100.} For all experiments, we used a batch size $b=1024$ with a latent dimension $d=128$. We optimized the ResNet18 with ADAM \cite{kingma2014adam} and a learning rate $\alpha=10^{-3}$.  We used a small weight decay $0.00005$. All images in a batch are transformed twice as in \cite{chen2020simCLR} through a set of transformations $\mathcal{T}$ implemented in PyTorch \cite{paszke2019pytorch} and including: Random Crop, Horizontal Flip, Color Jittering and Random Grayscale, following \cite{chen2020simCLR}. We used the standard testing split as defined in \cite{krizhevsky2009cifar} and we split the training set into training/validation with $n_{train}=40000$ and $n_{val}=10000$ randomly sub-sampled and stratified. The standard deviation in Fig. \ref{fig:cifar_linear_eval} is obtained by repeated all experiments 3 times with different random initialization. 

\paragraph{Brain MRI.} We followed the same experimental setting as \cite{dufumier2021yaware} for pre-training with a batch size $b=64$ and a learning rate $\alpha=10^{-4}$ reduced by $\gamma=0.9$ every 10 epochs. We trained the model for 50 epochs and evaluate the learned representation using a Logistic Regression implemented with scikit-learn \cite{pedregosa2011scikit}. For the transformations $\mathcal{T}$, we used random crop, random cutout, gaussian blur, gaussian noise and flip, as in \cite{dufumier2021yaware}. To obtain the standard deviation Fig. \ref{fig:linear_eval_biobd}, we performed 5-fold cross-validation on BIOBD.

\paragraph{BIOBD.} This dataset contains $N=662$ MRI scans acquired on 8 different sites with 356 HC and 306 patients with bipolar disorder, as described in \cite{dufumier2021yaware}.

\section{Proof of Lemma 1}

In order to prove the lemma 1, we first give a natural way of introducing the distribution $p_{pos}$, quantifying the similarity between positive samples $(x, x^+)$ and we then re-write the expectation under this distribution to derive the empirical estimator corresponding to the conditional alignment.

\paragraph{Discrete case.} As in \cite{saunshi2019theoretical}, we may define the notion of similarity through a distribution $p(c)$ defined over discrete latent classes $\mathcal{C}$ and a positive distribution $p^+(x, x^+)$: points $(x, x^+)$ are similar if and only if they share the same latent class $c$, \textit{i.e}, $x, x^+\sim p^+(x, x^+) \defeq \mathbb{E}_{p(c)}(p(x|c)p(x^+|c))$. In the fully supervised setting, the latent classes $\mathcal{C}$ are known and correspond to the available annotations $\mathcal{Y}$. 

\paragraph{General case.} In general, we do not know the latent classes $\mathcal{C}$ but we may replace them with the proxy labels $y\in \mathbb{R}^d$. Rewriting the positive distribution as:
\begin{equation*}
    p^+(x, x^+) =\int p(x|c)p(x^+|c^+)d\delta_c(c^+) p(c)dc
\end{equation*}
We may approximate this distribution by replacing the true latent class $c$ with the  \textit{proxy} label $y$ and the density $\delta_{y}(y^+)$ by an estimate $p_y^+(y^+)$ in the vicinity of $y$ (following \cite{chapelle2001vicinal}):
\begin{equation}
    \label{eq:pos_distrib}
    p_{pos}(x, x^+) = \int p(x|y) p(x^+|y^+) p^+_y(y^+)p(y)dydy^+
\end{equation}

Equipped with this distribution, and by remarking that the marginal distribution $p_{pos}(x) = \int p_{pos}(x, x^+)dx^+ = p(x)$, we can derive a lower bound of the mutual information between $x$ and $x^+$, $I(x, x^+)\defeq KL(p_{pos}(x, x^+)||p(x)p_{pos}(x^+))$ with the InfoNCE estimator \cite{gutmann2010infoNCE, oord2018representation}:
\begin{equation}
    I_{NCE}^y = \mathbb{E}_{\substack{(x, x^+)\sim p_{pos}(x, x^+)\\(x_i^-)_{i=1}^{N-1}\sim p(x^-)}}\log \frac{e^{f_\theta(x, x^+)}}{e^{f_\theta(x, x^+)}+\sum\limits_{i=1}^{N-1}e^{f_\theta(x,x_i^-)}} + \log(N)
\end{equation}

 where $f_\theta(x_1, x_2)=\frac{1}{\tau}f_\theta(x_1)^Tf_\theta(x_2)$ and $\tau>0$ is the temperature. 

\paragraph{y-Aware InfoNCE \cite{dufumier2021yaware}} We consider a Gaussian kernel function as a special case of density $p^+_y(y^+)=\frac{1}{Z_\sigma(y)}w_\sigma(y, y^+)p(y^+)$ (where $Z_\sigma(y)$ is a normalization constant and $w_\sigma$ is the Radial Basis Function kernel) to draw a connection with the recently introduced $y$-Aware InfoNCE loss \cite{dufumier2021yaware}. Let $g(x, x^+, (x_i^-)_{i=1}^{N-1})=\log \frac{e^{f_\theta(x, x^+)}}{e^{f_\theta(x, x^+)}+\sum\limits_{i=1}^{N-1}e^{f_\theta(x,x_i^-)}}$. We can re-write the previous $I_{NCE}^y$ estimator as:

\begin{align*}
    I_{NCE}^y &= \mathbb{E}_{\substack{(x, x^+)\sim p_{pos}(x, x^+)\\(x_i^-)_{i=1}^{N-1}\sim p(x^-)}}g(x, x^+, (x_i^-)_{i=1}^N) + \log(N)\\
    &= \mathbb{E}_{\substack{p(x, y)p(x^+|y^+)p_y^+(y^+)\\(x_i^-)_{i=1}^{N-1}\sim p(x^-)}}g(x, x^+,(x_i^-)_{i=1}^N) + \log(N)\\
    &= \mathbb{E}_{\substack{p(x, y)p(x^+, y^+)\\(x_i^-)_{i=1}^{N-1}\sim p(x^-)}} \frac{1}{Z_\sigma(y)}w_\sigma(y, y^+)g(x, x^+, (x_i^-)_{i=1}^N) + \log(N)\\
    &= \mathbb{E}_{\substack{p(x, y)p(x^+, y^+)\\(x_i^-)_{i=1}^{N-1}\sim p(x^-)}} \frac{1}{Z_\sigma(y)}w_\sigma(y, y^+)\log \frac{e^{f_\theta(x, x^+)}}{\frac{1}{N}\left(e^{f_\theta(x, x^+)}+\sum\limits_{i=1}^{N-1}e^{f_\theta(x,x_i^-)}\right)}
\end{align*}

This shows that $y$-aware InfoNCE loss $\mathcal{L}_{NCE}^y$ \cite{dufumier2021yaware} is the empirical estimator of $-I^{y}_{NCE}$. Then, we can decompose $I^y_{NCE}$ into 2 terms:

\begin{align*}
    I^y_{NCE} &= \mathbb{E}_{p_{pos}(x, x^+)}f_\theta(x, x^+) - \mathbb{E}_{\substack{(x, x^+)\sim p_{pos}\\(x_i^-)_{i=1}^N\sim p(x^-)}}\log \frac{1}{N}(e^{f_\theta(x, x^+)}+\sum\limits_{i=1}^{N-1}e^{f_\theta(x,x_i^-)})\\
\end{align*}

Finally, by the same proof describe in \cite{wang2020} (Appendix A.2), using the Strong Law of Large Numbers, the Continuous Mapping Theorem and the marginal $p_{pos}(x)=p(x)$, we have:
\begin{equation*}
   \lim_{N\rightarrow \infty}  \mathbb{E}_{\substack{(x, x^+)\sim p_{pos}\\(x_i^-)_{i=1}^N\sim p(x^-)}}\log \frac{1}{N}(e^{f_\theta(x, x^+)}+\sum\limits_{i=1}^{N-1}e^{f_\theta(x,x_i^-)}) = \mathbb{E}_{x\sim p(x)}\log \mathbb{E}_{x'\sim p(x')}(e^{f_\theta(x, x')})
\end{equation*}

This finally terminates the proof that:
\begin{equation*}
    \lim_{N\rightarrow \infty}\mathcal{L}_{NCE}^y = -\lim_{N\rightarrow\infty}I_{NCE}^y = -\mathbb{E}_{p_{pos}(x, x^+)}(f_\theta(x, x^+)) + \mathbb{E}_{x\sim p(x)}\log \mathbb{E}_{x'\sim p(x')}(e^{f_\theta(x, x')})
\end{equation*}


\end{document}